\pdfoutput=1

\documentclass[11pt]{article}

\usepackage[preprint]{acl}

\usepackage{times}
\usepackage{latexsym}

\usepackage[T1]{fontenc}

\usepackage[utf8]{inputenc}
\usepackage{tabularx}
\usepackage{microtype}
\usepackage{inconsolata}
\usepackage{graphicx}
\usepackage{booktabs}
\usepackage{amsmath}
\usepackage{multirow}
\usepackage{array}
\usepackage{xspace}
\usepackage{xcolor}
\usepackage[table]{xcolor}
\newcommand{\method}{SelfMem\xspace}
\newcommand{\beam}{BEAM\xspace}

\title{SelfMem: Self-Optimizing Memory for AI Agents}

\author{ 
    \textbf{Shu Yang\textsuperscript{1}},
    \textbf{Junchao Wu\textsuperscript{2}},
    \textbf{Derek F. Wong\textsuperscript{2}},
    \textbf{Di Wang\textsuperscript{1}†} \\[4pt]
    \textsuperscript{1}PRADA Lab, King Abdullah University of Science and Technology \\
    \textsuperscript{2} NLP2CT Lab, University of Macau
}

\begin{document}
\maketitle
\def\thefootnote{†}\footnotetext{Corresponding Author}
\begin{abstract}
While current AI agents support increasingly long context windows, tool use, and skill execution for long-horizon tasks, they still require memory systems to effectively leverage historical experience. Existing memory frameworks typically rely on \textit{fixed} storage, retrieval, and summarization mechanisms, which can be rigid across different tasks and often require manual tuning. To address this limitation, we propose \method, a self-optimizing memory framework. Inspired by prior work on self-improving AI, we follow the principle of ``teaching an agent to fish rather than giving it a fish.'' Instead of forcing the model to follow a predefined memory strategy or format, \method provides an environment with memory tools and feedback signals that allow the agent to explore, evaluate, and refine its own memory strategy.  Our results show that \method consistently outperforms retrieval, compression, and agent-memory baselines on \beam across conversation scales from 100K to 1M tokens. 
Compared with the strongest baseline, \method improves the official score by 48.7\%, 40.8\%, and 41.9\% at 100K, 500K, and 1M, respectively. Further question-type analysis shows broad robustness across diverse memory demands, and our optimization study shows that model-guided strategy refinement further improves performance. 
\end{abstract}

\section{Introduction}
With the growing ability to process longer contexts~\citep{bai2024longbench}, perform multi-step planning~\citep{li2025agentoriented}, use external tools~\citep{schick2023toolformer}, and incorporate environmental feedback~\citep{yao2023react,shinn2023reflexion,zhou2024language}, AI agents have demonstrated strong capabilities across increasingly diverse long-horizon tasks~\citep{liu2024agentbench,merrill2026terminal}. However, these advances also place greater demands on agents' ability to remember user preferences, project backgrounds, and important factual details over extended interactions, making effective memory-system support necessary~\citep{liu2024lost,packer2023memgpt,zhong2024memorybank,tavakoli2025beyond}.

\begin{figure}[t]
    \centering
    \includegraphics[width=\linewidth]{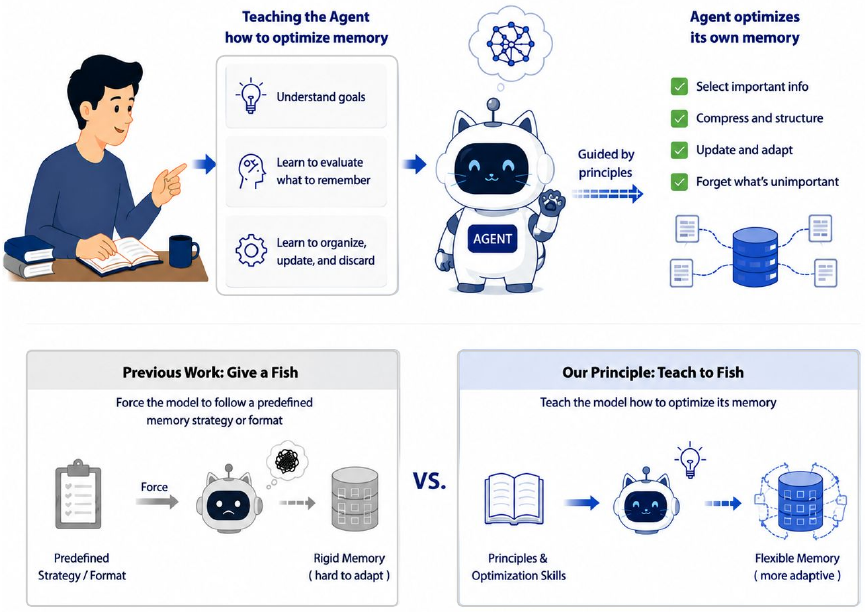}
    \caption{``Teach agent to fish rather than give it a fish'': guiding agents to self-optimize memory by learning
    principles and trade-offs, rather than relying on rigid,
    predefined memory strategies.}
    \label{fig:memfish_intro}
\end{figure}

\begin{figure*}[t]
    \centering  \includegraphics[width=\textwidth]{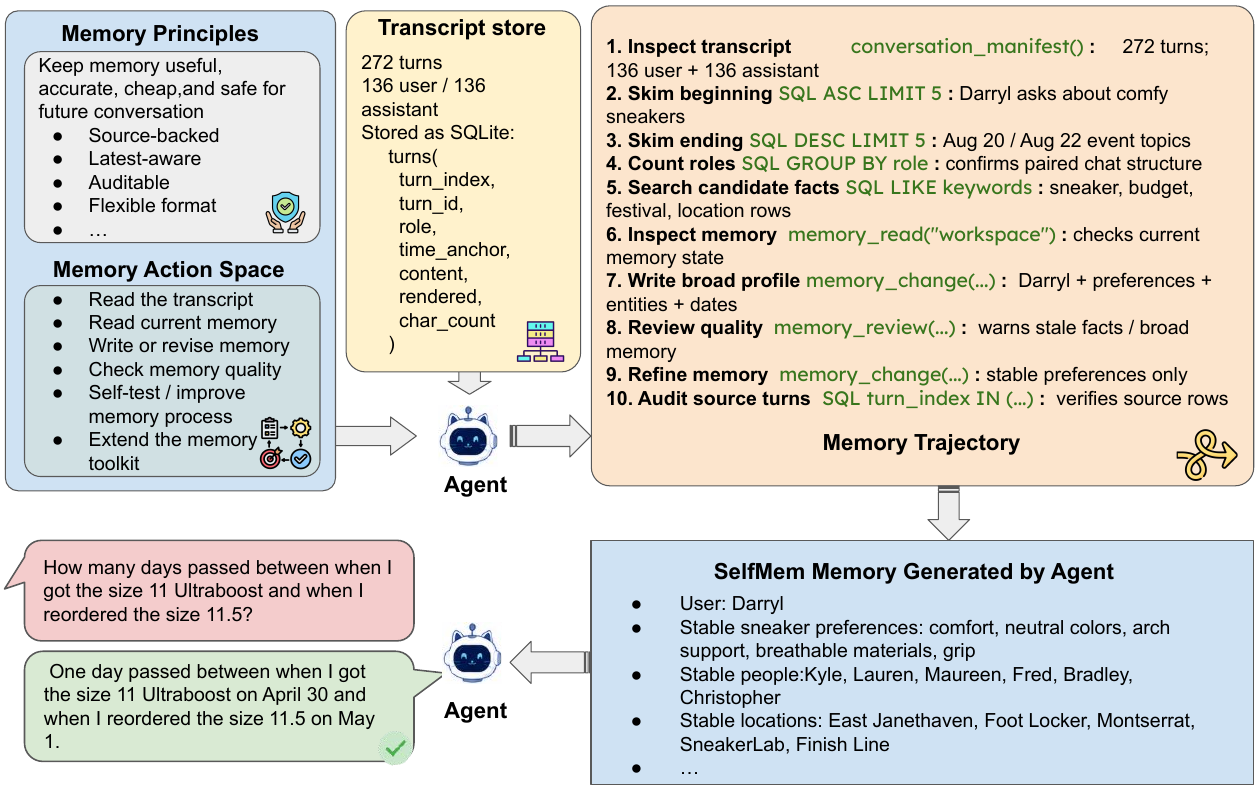}
    \caption{Overview of \method. The raw transcript is kept as an immutable source of truth, while the agent is given a memory action space for inspecting the transcript, writing and revising memory, and reviewing memory quality.}
    \label{fig:selfmem-overview}
\end{figure*}

Existing agent-memory methods, such as LoCoMo, retrieval-augmented generation (RAG), MemGPT, and MemoryBank, typically rely on manually specified strategies to construct, update, retrieve, and present historical knowledge to the model~\citep{maharana2024locomo,packer2023memgpt,zhong2024memorybank}. Although these methods improve agents' ability to use past conversation histories, their memory structures and operations are still largely predefined and limit their ability in broader tasks. A-Mem ~\citep{xu2025amem} takes a more agentic step by allowing agents to write, index, retrieve, and link memories themselves, following the Zettelkasten principle of maintaining an evolving knowledge network. 
However, the core challenge remains: different tasks and conversation histories require different memory behaviors, yet it is still unclear how agents should adapt their memory strategies to task demands and environmental feedback. As illustrated in Figure~\ref{fig:memfish_intro}, this motivates our method, \method: instead of giving agents a fixed memory mechanism, we teach them how to manage memory adaptively by exposing memory-management decisions to the agents themselves.

As shown in Figure~\ref{fig:selfmem-overview}, \method turns memory management into an agent-controlled optimization process. Rather than prescribing a fixed storage, update, or retrieval rule, \method exposes memory tools together with feedback signals such as response performance, token counts, cost, and cache reuse. Using these signals, the agent decides what to keep in memory, what to compress, what to update, and what to leave for transcript retrieval. Empirically, our method yields a stronger quality--cost trade-off than fixed memory pipelines.

Our results show that \method delivers robust memory performance on \beam across long conversation histories, diverse question types, and all evaluated token scales. \method achieves the highest official score and Pass$_{0.5}$ at 100K, 500K, and 1M tokens. Compared with the strongest non-\method baseline, it improves the official score by 0.165, 0.141, and 0.134, respectively, and increases Pass$_{0.5}$ by 17.0, 14.9, and 14.3 percentage points. Beyond aggregate performance, question-type analysis shows that \method remains broadly robust across different memory demands, achieving the best score on 9/10 question types at 100K, 8/10 at 500K, and 7/10 at 1M. Our optimization and generalization analysis further shows that model-guided strategy refinement can also improve memory behavior, suggesting that \method exposes a useful optimization surface for long-horizon memory management.

\section{Related Work}
\label{sec:related}

\paragraph{Agent Memory}
Recent work on agent memory equips LLM agents with external memory states for long-running interaction and personalization. Generative Agents introduced a memory-stream architecture in which agents record observations, retrieve relevant experiences, and synthesize higher-level reflections for future planning \citep{park2023generative}. MemoryBank extends this direction to long-term dialogue by maintaining persistent memories and user portraits that can be recalled, updated, and reinforced over repeated interactions \citep{zhong2024memorybank}. MemGPT formulates long-context interaction as virtual context management, where an agent moves information between core/working memory and archival memory through tool-mediated memory operations \citep{packer2023memgpt}. ReadAgent studies agentic reading over very long inputs by constructing gist memories and selectively looking up original passages when detailed evidence is needed \citep{lee2024readagent}. A parallel line of study examines how to evaluate and structure long-term conversational memory. LoCoMo introduces very long-term multi-session dialogues and evaluates agents on question answering, event summarization, and multimodal dialogue generation over extended histories \citep{maharana2024locomo}. LongMemEval further decomposes long-term memory evaluation into abilities such as information extraction, multi-session reasoning, temporal reasoning, knowledge updates, and abstention \citep{wu2024longmemeval}. More recent memory systems move toward structured and production-oriented memory construction. Mem0 extracts, consolidates, and retrieves salient conversational facts, with an enhanced graph-based variant for relational memory \citep{chhikara2025mem0}. A-Mem takes a more agentic step by dynamically writing, indexing, linking, and evolving memories following Zettelkasten-style organization \citep{xu2025amem}. As summarized in Appendix~\ref{app:implementation}, Table~\ref{tab:memory-components}, these methods differ in their memory units, retrieval mechanisms, and update behaviors, but they still largely instantiate a predefined memory architecture or a fixed set of memory operations. 

\paragraph{Self-improving Agent}
A complementary line of work shows that language models and agents can improve their behavior through feedback, reflection, and search, often without updating the underlying model parameters. Self-Refine uses the same model to generate feedback and iteratively refine its own outputs \citep{madaan2023selfrefine}. Reflexion converts task feedback into verbal reflections stored in episodic memory, allowing agents to improve across future trials through language rather than weight updates \citep{shinn2023reflexion}. Voyager demonstrates that an embodied agent can use environmental feedback, execution errors, and self-verification to build an expanding library of reusable skills \citep{wang2023voyager}. LATS further combines reasoning, acting, planning, and environmental feedback through tree search, enabling agents to explore and evaluate alternative action trajectories before committing to a decision \citep{zhou2024language}. Self-RAG extends this idea to retrieval-augmented generation by learning when to retrieve, generate, and critique evidence through self-reflection \citep{asai2024selfrag}. Beyond improving a single output or action trajectory, recent work on natural-language optimization suggests that models can also search over strategies themselves. OPRO treats an LLM as a black-box optimizer that proposes new candidate instructions from previous solutions and their scores \citep{yang2024large}. \citet{pryzant2023automatic} uses natural-language gradients, beam search, and bandit selection to automatically improve prompts. Promptbreeder evolves both task prompts and mutation prompts through fitness feedback, showing that hand-crafted prompt strategies can be improved through iterative self-referential search \citep{fernando2024promptbreeder}. These studies motivate our hypothesis that memory behavior can also be optimized in language: given appropriate feedback signals and an action space, the agent may discover memory strategies that are better adapted to the current task than manually specified rules.

\section{\method: A Self-Optimizing Memory Framework}
\label{sec:method}
\subsection{Overview}

As shown in Figure~\ref{fig:selfmem-overview}, \method turns memory management into a model-controlled optimization process. The framework keeps the original logged conversation history (raw transcript) as an immutable source of truth, while exposing a memory workspace and a set of memory-management tools to the agent. Instead of asking the agent to fill a fixed profile schema or follow a predefined summarization rule, \method allows the agent to inspect transcript evidence, decide what memory is worth writing, organize memory in a task-sensitive format, review its own memory quality, and revise memory when diagnostics indicate risk. The key design goal is to separate \textit{framework constraints} from \textit{memory strategy}. The framework provides human-designed procedural principles, the available memory tools, feedback channels, and audit constraints. The agent then determines the concrete memory behavior: which transcript turns or retrieved snippets to inspect, what information to store, what to compress or update, what to preserve exactly, and what to leave recoverable from the original transcript.

\subsection{Problem Definition}

We consider memory management for agents that operate over a fixed interaction history. Let
\[
H = (t_1, t_2, \ldots, t_T)
\]
denote the chronological record of previous user--agent turns, where each turn contains the speaker role, time anchor, content, and rendering metadata. Given a future query \(q\), the agent must produce an answer \(y\) by using information from \(H\). A simple solution is to place the entire history into the model context. However, this becomes expensive or infeasible when the interaction exceeds the context window. Another common solution is to construct a fixed summary or memory schema, but such representations may omit task-critical details, preserve stale facts, or fail to match the needs of different tasks. We therefore formulate memory management as the construction of a compact memory workspace \(M^\star\) from the original interaction history, while keeping the transcript available for exact factual grounding:
\[
\begin{aligned}
M^\star &= \Pi_{\mathrm{agent}}(H; \mathcal{A}, \mathcal{F}), \\
E_q &= \mathrm{Retrieve}(q, M^\star, H), \\
y &= \mathrm{LLM}(q, M^\star, E_q).
\end{aligned}
\]
Here, \(\mathcal{A}\) is the memory action space available to the agent, \(\mathcal{F}\) is the feedback returned by the environment, and \(E_q\) denotes the transcript snippets or turns retrieved to support the query. Existing memory systems usually instantiate \(\Pi\) with hand-designed storage formats, retrieval rules, summarization policies, or update operations. In contrast, \method exposes this memory-management procedure to the agent itself: the agent decides which parts of the transcript to inspect, what information to store, how to organize it, and when to revise it based on tool feedback and task demands.

\subsection{Transcript Store}

The transcript store is the source of truth in \method. Rather than exposing hidden summaries or pre-extracted facts, \method stores the logged interaction history as a read-only database. In our implementation, each transcript is represented as a SQLite table:

\begin{center}
\small
\begin{tabular}{@{}l@{}}
\texttt{turns(} \\
\quad \texttt{turn\_index, turn\_id, role, time\_anchor,} \\
\quad \texttt{content, rendered, char\_count} \\
\texttt{)}
\end{tabular}
\end{center}

The memory agent can inspect this transcript only through read-only tools:
\[
o_i^{H} = \mathrm{Read}_{H}(a_i),
\]
where \(a_i\) is a model-selected read action and \(o_i^{H}\) is the returned transcript observation. The transcript interface includes \texttt{conversation\_manifest}, which reports schema and transcript statistics, and \texttt{conversation\_sql\_query}, which allows the agent to issue read-only SQL queries over the turns table. This design lets the agent decide which parts of the history are worth inspecting, rather than forcing a fixed chunking or summarization policy over the entire conversation.

\subsection{Memory Workspace and Action Space}

We provide a model-managed memory workspace \(M_i\). The workspace can contain any structure the agent judges useful, such as a user profile, preference list, project-state note, timeline, or even compact policy. The agent interacts with the workspace through a model-selectable action space:
\[
\mathcal{A}
=
\mathcal{A}_{\mathrm{read}}
\cup
\mathcal{A}_{\mathrm{write}}
\cup
\mathcal{A}_{\mathrm{review}}
\cup
\mathcal{A}_{\mathrm{probe}} .
\]
At each step, the agent selects both a tool and its arguments:
\[
a_i =
\mathrm{Select}_{\mathrm{LLM}}
(P_{\mathrm{mem}}, I, M_{i-1}, o_{<i}, d_{<i}),
\]
where \(P_{\mathrm{mem}}\) is the memory-management prompt, \(I\) denotes the procedural memory principles, \(M_{i-1}\) is the current workspace, \(o_{<i}\) are previous tool observations, and \(d_{<i}\) are previous diagnostic signals.

The main tool families are:

\textit{Transcript-read tools.}
These tools inspect the immutable transcript. The agent can obtain the transcript manifest, run read-only SQL queries, search for entities or events, check recent turns, and audit source turns before or after writing memory.
\textit{Memory-read tools.}
The agent can inspect the current workspace using them. This allows it to avoid duplicate writes, detect stale memory, and decide whether a new update should add, replace, merge, demote, or delete existing content.
\textit{Memory-write tools.}
The agent modifies memory through these tools. Each change contains an operation type, target, content, rationale, and metadata. Supported operation styles include adding memory, replacing memory, merging duplicates, refining broad summaries, archiving stale information, recording exact facts, creating timelines, and adding retrieval hints.
\textit{Memory-review tools.}
The agent can call the tools to diagnose an existing or proposed memory without mutating it. Review feedback can flag missing source support, stale exact facts, contradictions, overly broad summaries, and retrievability risk.
\textit{Self-test tools.}
Optionally, the agent can create internal memory probes from the observed transcript and workspace. These probes test whether the memory organization supports plausible future questions, without using hidden benchmark answers, rubrics, or per-question judge labels.

\subsection{Feedback Signals}

The feedback \(\mathcal{F}\) returned by \method is tool and environment feedback rather than a hidden supervision signal. At memory-construction time, the agent does not receive gold answers, benchmark rubrics, held-out labels, or per-question judge decisions. Instead, feedback comes from observable interactions with the transcript, memory workspace, and diagnostics:
\[
\mathcal{F}_i =
\{o_i^{H}, o_i^{M}, d_i^{\mathrm{write}}, d_i^{\mathrm{review}}, u_i\},
\]
where \(o_i^{H}\) is transcript-read feedback, \(o_i^{M}\) is memory-read feedback, \(d_i^{\mathrm{write}}\) is write-time diagnostic feedback, \(d_i^{\mathrm{review}}\) is review feedback, and \(u_i\) denotes optional operational signals such as token usage, latency, cache reuse, request count, and estimated cost. These signals are deliberately not collapsed into a single scalar reward. 
Memory management involves trade-offs, such as storing more information can improve recall but increase context cost; compressing memory can reduce latency but lose exact details. \method lets the agent reason about these trade-offs in language and revise its memory behavior accordingly.

\subsection{Memory Construction and Review}

Memory construction proceeds as a model-controlled tool-use loop. The workspace is initialized as
\[
M_0 = \emptyset .
\]
At step \(i\), the agent observes the current workspace and previous tool outputs, then selects an action \(a_i\). Read and review actions return observations or diagnostics, while write actions produce an update \(\Delta M_i\) that is applied to the workspace:
\[
M_i =
\begin{cases}
\mathrm{Apply}(M_{i-1}, \Delta M_i), & a_i \in \mathcal{A}_{\mathrm{write}}, \\
M_{i-1}, & \text{otherwise}.
\end{cases}
\]

Each write operation records its target, content, rationale, and metadata, making the memory process auditable. Since memory writing is not treated as one-shot summarization, the agent can review a draft or existing memory:
\[
d_i = \mathrm{Review}(M_i, f_i),
\]
where \(f_i\) specifies the review focus, such as source support, exact facts, stale values, contradictions, or retrievability. The agent can then revise the workspace through another memory update. This creates an inspect--write--review--revise loop in which the agent improves the memory organization itself, rather than simply producing a final answer.

\subsection{Memory-Conditioned Answering}

After construction, the final workspace \(M^\star\) is used as a compact guide for future answering. Given a query \(q\), \method retrieves relevant transcript turns or snippets \(E_q\) and provides them together with the memory workspace to the answer model:
\[
\begin{aligned}
E_q &= \mathrm{Retrieve}(q, M^\star, H), \\
y &= \mathrm{LLM}(q, M^\star, E_q).
\end{aligned}
\]

In \method, the memory workspace is not populated by a hand-written rule that specifies what must be saved. Instead, the agent decides what information is worth storing based on the transcript, available memory tools, and feedback signals. It may keep durable information such as user preferences, recurring entities, and project context, while leaving exact or frequently changing details, such as dates, counts, prices, and updated decisions, to be retrieved from relevant transcript turns when needed.

Overall, \method shifts memory from a fixed storage pipeline to a model-guided process: the framework provides the transcript store, tools, feedback channels, and constraints, while the agent decides the concrete memory actions and organization.

\section{Experiments}
\label{sec:experiments}

\subsection{Experimental Setup}
\label{sec:exp-setup}

\noindent\textbf{Benchmark and evaluation protocol.}
We evaluate on the original \beam benchmark~\citep{tavakoli2025beyond}, which provides \textit{long conversation histories} at 100K, 500K, and 1M token scales, together with 100 conversations and 2,000 probing questions that test memory over extended interactions, covering information extraction, temporal reasoning, multi-session reasoning, and event ordering. Questions are asked sequentially, with previous model-generated question--answer turns appended to later prompts. This simulates a multi-turn probing session, enables prompt-cache reuse, and is applied consistently to all compared methods.

\noindent\textbf{Baselines.}
Following the methods reviewed in Section~\ref{sec:related}, we compare \method with representative memory and context-management baselines. These include RAG, which chunks, embeds, and retrieves conversation history~\citep{lewis2020retrieval}; full context, which directly places the raw conversation in the prompt when feasible; and compression, which compacts history when it approaches the context limit. We also include LoCoMo, MemoryBank, ReadAgent, MemGPT, A-Mem, and Mem0, covering persistent conversational memory, agentic reading, hierarchical memory, linked note-style memory, and production-oriented memory extraction~\citep{maharana2024locomo,zhong2024memorybank,lee2024readagent,packer2023memgpt,xu2025amem,chhikara2025mem0}. All baselines use the same conversations, questions, answer model, and official judge; they differ only in how the conversation history is transformed into the context given to the answer model. Appendix~\ref{app:implementation} provides implementation details.

\noindent\textbf{Model and decoding configuration.}
All LLM-generation calls in memory construction, and answer generation use GPT-5.4-nano, and for BEAM score we use GPT-5.4-mini as the judge model. We use deterministic decoding with temperature \(0\) for answer generation and LLM-based judging. Full-context baseline prompting is evaluated only when the raw conversation fits within the model and API limits; it is marked unsupported at larger scales when the original history exceeds those limits.

\noindent\textbf{Evaluation metrics.}
Following the \beam evaluation protocol, \textit{Score }is the mean official question score across all evaluated probing questions: nugget-based questions use the official LLM judge score, while event-ordering questions use normalized Kendall correlation, \(\tau_{\mathrm{norm}}=(\tau_b+1)/2\). Pass$_{0.5}$, is a thresholded pass rate: the percentage of questions whose official score is at least \(0.5\). \textit{Cost} is token-computed API cost, calculated from recorded request-level token usage and the model pricing table. \textit{Cache hit} is the ratio of cached input tokens to total input tokens. \textit{LLM Req.} and \textit{Embed Req.} count LLM and embedding requests, respectively.

\begin{table*}[t]
\centering
\tiny
\setlength{\tabcolsep}{3.5pt}
\renewcommand{\arraystretch}{0.88}
\resizebox{\textwidth}{!}{%
\begin{tabular}{c l c c c c c c}
\toprule
\textbf{Scale} & \textbf{Method}
& \textbf{Pass$_{0.5}$, $\uparrow$}
& \textbf{Score $\uparrow$}
& \textbf{Cost $\downarrow$}
& \textbf{Cache}
& \textbf{LLM Req.}
& \textbf{Embed Req.} \\
\midrule

\textbf{100K} & RAG & \underline{42.00} & \underline{0.339} & \textbf{\$0.551} & 1.7\% & 1{,}548 & 5{,}656 \\
& Full context & 25.80 & 0.205 & \$2.427 & 87.7\% & 1{,}559 & 0 \\
& Compression & 22.50 & 0.170 & \$2.830 & 85.0\% & 1{,}726 & 0 \\
& LoCoMo & 36.50 & 0.306 & \$2.904 & 1.1\% & 1{,}587 & 580 \\
& ReadAgent & 37.00 & 0.314 & \$2.940 & 1.0\% & 1{,}600 & 490 \\
& MemoryBank & 34.50 & 0.284 & \underline{\$0.609} & 2.1\% & 1{,}602 & 6{,}212 \\
& MemGPT & 36.20 & 0.291 & \$6.434 & 1.0\% & 1{,}582 & 450 \\
& A-Mem & 35.20 & 0.302 & \$3.116 & 0.7\% & 1{,}606 & 580 \\
& Mem0 & 35.80 & 0.282 & \$1.697 & 34.1\% & 2{,}160 & 2{,}912 \\
\rowcolor{gray!12}
& \textbf{\method (Ours)} & \textbf{59.00} & \textbf{0.504} & \$0.984 & 36.2\% & 1{,}932 & 0 \\

\midrule
\textbf{500K} & RAG & \underline{42.10} & \underline{0.346} & \textbf{\$1.258} & 0.6\% & 2{,}821 & 37{,}011 \\
& Full context & -- & -- & -- & -- & -- & -- \\
& Compression & 25.30 & 0.195 & \$8.144 & 71.4\% & 4{,}066 & 0 \\
& LoCoMo & 36.90 & 0.302 & \$8.156 & 0.6\% & 2{,}812 & 1{,}400 \\
& ReadAgent & 35.90 & 0.286 & \$7.926 & 0.8\% & 2{,}799 & 1{,}050 \\
& MemoryBank & 33.90 & 0.276 & \underline{\$1.335} & 0.5\% & 2{,}799 & 39{,}897 \\
& MemGPT & 30.70 & 0.249 & \$11.917 & 1.2\% & 2{,}799 & 980 \\
& A-Mem & 34.90 & 0.281 & \$6.500 & 0.6\% & 2{,}812 & 1{,}400 \\
& Mem0 & 36.40 & 0.303 & \$9.366 & 36.4\% & 6{,}846 & 13{,}898 \\
\rowcolor{gray!12}
& \textbf{\method (Ours)} & \textbf{57.00} & \textbf{0.487} & \$1.828 & 33.8\% & 3{,}559 & 0 \\

\midrule
\textbf{1M} & RAG & \underline{38.30} & \underline{0.320} & \textbf{\$1.843} & 0.4\% & 3{,}468 & 74{,}153 \\
& Full context & -- & -- & -- & -- & -- & -- \\
& Compression & 25.70 & 0.210 & \$15.033 & 57.3\% & 6{,}376 & 0 \\
& LoCoMo & 29.30 & 0.261 & \$12.275 & 0.9\% & 3{,}662 & 1{,}421 \\
& ReadAgent & 30.70 & 0.268 & \$12.232 & 0.6\% & 3{,}618 & 1{,}061 \\
& MemoryBank & 31.40 & 0.266 & \underline{\$1.934} & 0.4\% & 3{,}499 & 76{,}903 \\
& MemGPT & 25.70 & 0.221 & \$12.264 & 0.8\% & 3{,}609 & 989 \\
& A-Mem & 31.10 & 0.265 & \$8.480 & 0.7\% & 3{,}588 & 1{,}421 \\
& Mem0 & 34.90 & 0.292 & \$18.830 & 35.9\% & 11{,}356 & 25{,}779 \\
\rowcolor{gray!12}
& \textbf{\method (Ours)} & \textbf{52.57} & \textbf{0.454} & \$2.004 & 31.1\% & 4{,}040 & 0 \\

\bottomrule
\end{tabular}%
}
\caption{Main \beam results across all conversations. Bold and underlined values mark the best and second-best results within each scale for Pass$_{0.5}$, Score, and Cost. Cache hit and request counts are reported for efficiency but not highlighted. Cost includes answer-generation and judge calls; token-count requests are excluded.}
\label{tab:beam-scores}
\end{table*}

\subsection{Results}
\label{sec:results}

\noindent\textbf{\method consistently achieves the best memory performance.} Table~\ref{tab:beam-scores} reports the main \beam results across 100K, 500K, and 1M token scales. \method achieves the best Score and Pass$_{0.5}$ at all scales, indicating that its advantage is not limited to a particular context length. Compared with fixed retrieval, compression, and structured memory baselines, \method maintains stronger performance as the conversation history grows longer. This suggests that agent-controlled memory construction is better suited to long-horizon settings, where relevant information is sparse, distributed across many turns, and often mixed with outdated or less useful details. The strongest baseline is consistently RAG, which shows that simple retrieval remains a competitive strategy for long conversations. However, RAG only retrieves chunks at query time and does not build an adaptive memory representation. In contrast, \method allows the agent to decide what should be stored, revised, compressed, or left for later transcript lookup. This leads to large and stable gains over the best non-\method baseline, with Score improvements of 0.165, 0.141, and 0.134 at 100K, 500K, and 1M, respectively. The persistence of these gains at 1M tokens shows that \method scales more robustly than manually designed memory pipelines.

\noindent\textbf{Better accuracy does not require expensive memory pipelines.}
Table~\ref{tab:beam-scores} also shows that higher accuracy is not simply the result of using heavier memory infrastructure. Several baselines introduce substantial additional cost through repeated LLM calls, embedding-based ingestion, or complex memory updates, yet still underperform \method. For example, at 1M tokens, Mem0 costs \$18.830 and uses 25{,}779 embedding requests, but reaches only 0.292 Score; MemGPT and A-Mem also incur much higher costs than \method while achieving lower accuracy. In contrast, \method obtains the best 1M Score of 0.454 with a cost of only \$2.004 and zero embedding requests. Although RAG remains the cheapest method, its lower scores across all scales indicate that inexpensive retrieval alone is insufficient for robust long-horizon memory. These results suggest that \method improves the quality--cost trade-off by letting the agent selectively construct and revise useful memory, rather than relying on exhaustive retrieval, ingestion, or fixed memory pipelines.

\begin{figure*}[t]
\centering
\includegraphics[width=\textwidth]{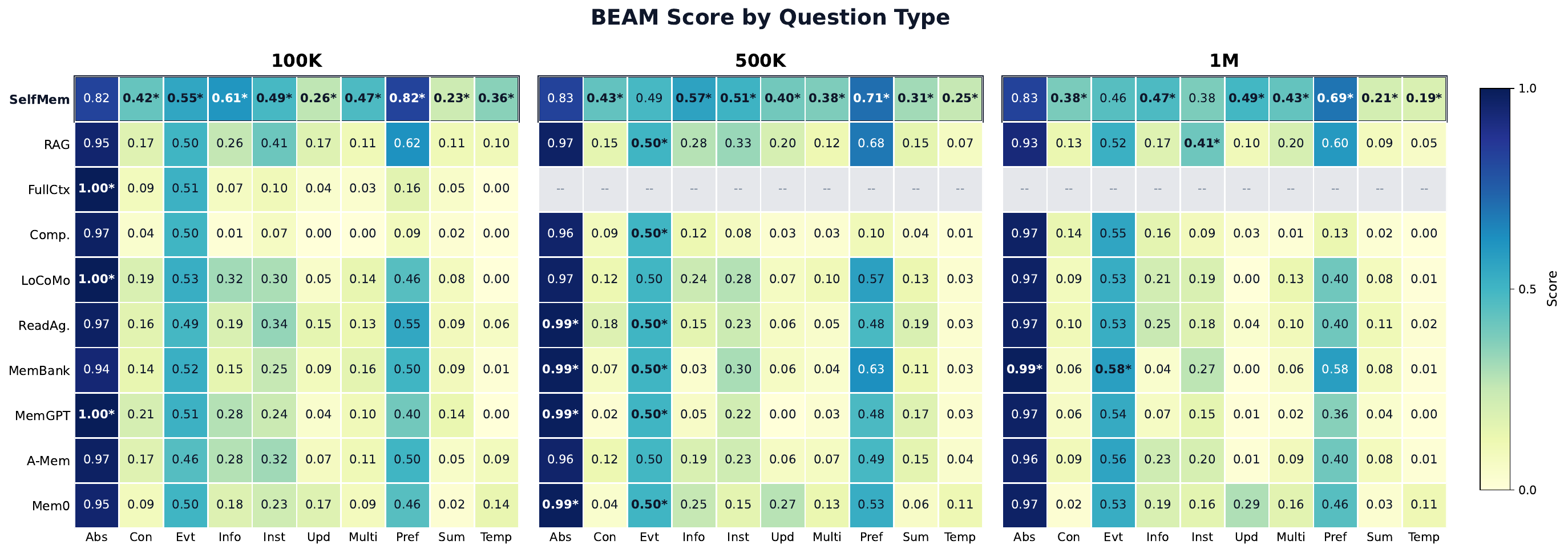}
\caption{Question-type analysis on BEAM. Each cell reports the official BEAM score averaged over evaluated conversations for a method, context scale, and question type. Asterisks mark the best method within
each scale and question type.}
\label{fig:beam-qtype-heatmap}
\end{figure*}

\noindent\textbf{\method is more robust across question types.}
Figure~\ref{fig:beam-qtype-heatmap} provides a finer-grained breakdown by \beam question type. \method performs well across diverse memory demands rather than improving only one favorable category. It achieves the best score on most question types at every scale, including 9/10 types at 100K, 8/10 at 500K, and 7/10 at 1M. The strongest gains appear in information extraction, knowledge update, multi-session reasoning, preference following, summarization, and temporal reasoning, where answers require selecting and organizing sparse information from long conversation histories. This pattern supports the core design of \method: allowing the agent to construct and revise memory helps it adapt to different needs.

\begin{figure}[t]
\centering
\includegraphics[width=0.5\textwidth]{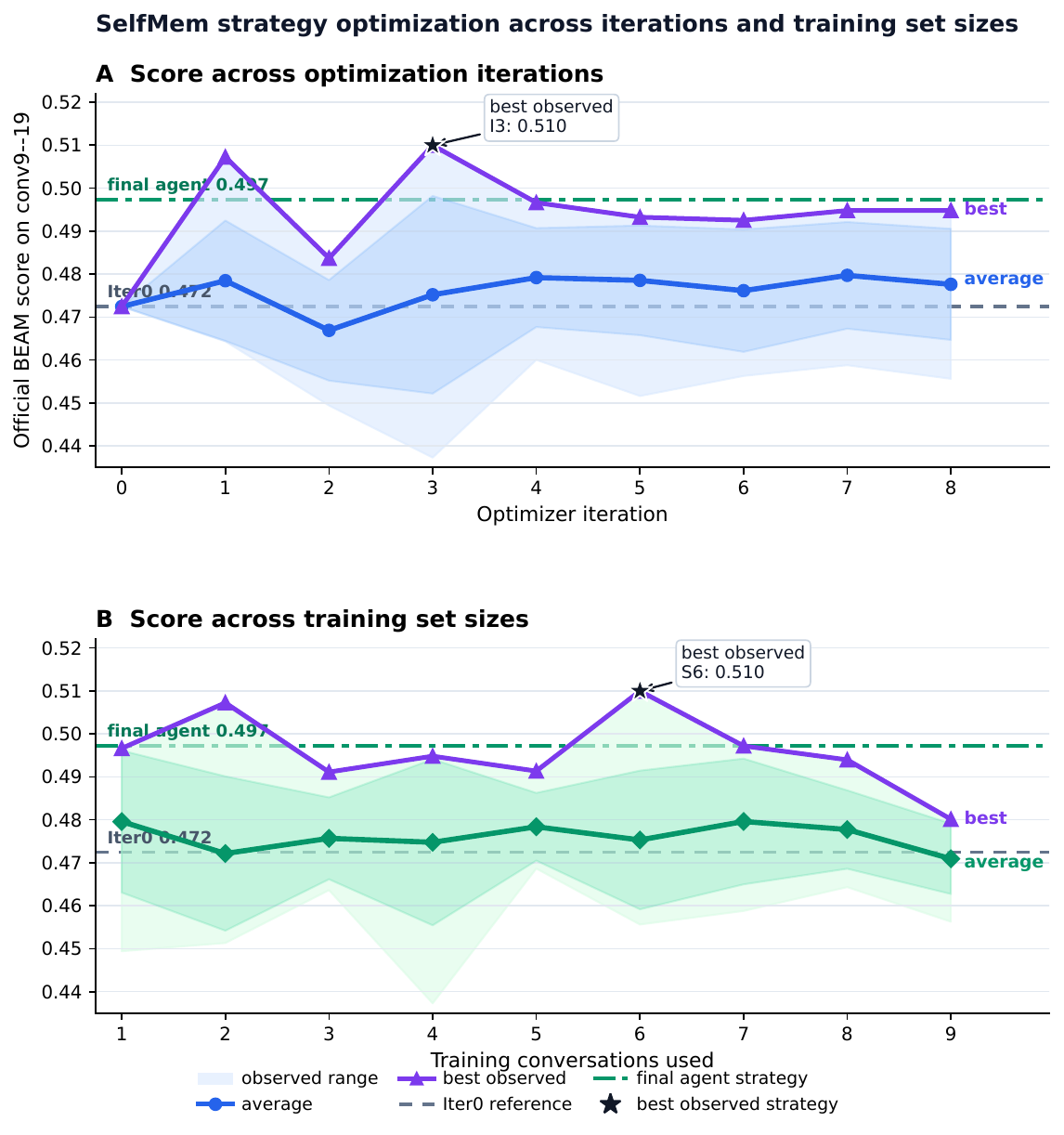}
\caption{
SelfMem strategy optimization across refinement iterations and training set sizes.
The upper panel reports held-out BEAM scores across optimizer iterations, while the lower panel groups refinement outcomes by the number of training conversations used.
}
\label{fig:selfmem-optimization-lines}
\end{figure}

\section{Optimization and Generalization Analysis}
We next evaluate whether \method can further improve its memory behavior through iterative strategy refinement. Starting from the default setting, the agent repeatedly proposes revised memory-strategy notes using only aggregate feedback from training conversations, including training scores and memory/tool diagnostics. Candidate strategies are generated from the training split conversation 0-8 and evaluated on a separate held-out test split conversation 9-19 in BEAM~\cite{tavakoli2025beyond} dataset. Iteration 0 corresponds to the default \method setting without refinement, while later iterations use model-produced strategy notes. Held-out test scores are used only for final analysis and are never provided to the agent during strategy generation. Further details about the split construction, refinement procedure, and prompts are provided in Appendix~\ref{app:selfmem-strategy} and Appendix~\ref{app:prompt-templates}.

\noindent\textbf{Results.}
Figure~\ref{fig:selfmem-optimization-lines} shows how to read the refinement process from two complementary views. The upper panel groups results by refinement round, showing whether repeated strategy revision improves memory behavior over time. The lower panel groups results by the number of training conversations, showing whether using more training feedback leads to better held-out performance. In both panels, the gray band indicates the observed performance range, the average line shows the typical strategy quality, and the best-strategy line highlights the strongest memory policy discovered during the search. This layout separates average stability from peak improvement, making it clear that refinement can discover better strategies even when the average trend is not strictly increasing. Starting from the default no-refinement strategy, which achieves a held-out score of 0.472, the final agent-synthesized strategy improves to 0.497, and the best strategy discovered during refinement reaches 0.510. The strongest result appears at an intermediate refinement stage and with an intermediate amount of training feedback, rather than at the largest iteration or largest training set. This suggests that the gain does not come from simply running more refinement rounds or adding more examples. Instead, \method benefits when the agent discovers a more robust procedural strategy for deciding what to store, revise, compress, and retrieve. Overall, Figure~\ref{fig:selfmem-optimization-lines} shows that \method exposes a useful refinement space for memory management, where better memory behavior can be learned from aggregate feedback and memory-tool diagnostics. 

\section{Conclusion}

We introduced \method, a self-optimizing memory framework for AI agents. Rather than placing the agent inside a fixed retrieval, compression, or memory-update pipeline, \method exposes memory tools and feedback signals that allow the agent to decide what to store, revise, compress, and retrieve. \method achieves the best performance in BEAM at 100K, 500K, and 1M token scales, improving over the strongest non-\method baseline by 0.165, 0.141, and 0.134 Score points, respectively. These gains are achieved with substantially lower cost than heavier memory frameworks such as MemGPT, A-Mem, and Mem0. Also, our strategy refinement analysis further shows that memory behavior can be improved through iterative strategy updates.  Together, these results indicate that providing the model with an environment for learning and exploring memory strategies is more effective than directly prescribing a fixed memory pattern.
\newpage

\section*{Limitations}
This work focuses on evaluating the core hypothesis of \method: providing agents with an environment to explore and refine memory strategies can be more effective than prescribing a fixed memory pattern. Our experiments are conducted on \beam with a fixed answer and judging protocol, so future work should validate the approach on additional long-horizon benchmarks, model families, and deployment settings. The iterative strategy-refinement study is currently performed at the 100K scale; extending the same refinement procedure to 500K and 1M histories would further test its scalability. In addition, our baselines are implemented in a shared harness to ensure fair comparison, but highly specialized production implementations may show different efficiency profiles. These limitations mainly concern evaluation breadth and engineering scope. They do not change the main finding: under the same benchmark, model, and judging protocol, \method consistently improves long-horizon memory performance while avoiding heavier fixed memory pipelines.

\bibliography{custom}

\appendix
\newpage
\section{Baseline implementation details}
\label{app:implementation}

\begin{table*}[t]
\centering
\scriptsize
\setlength{\tabcolsep}{3.2pt}
\renewcommand{\arraystretch}{1.18}
\newcolumntype{Y}{>{\raggedright\arraybackslash}X}

\begin{tabularx}{\textwidth}{@{}p{1.55cm}p{2.15cm}p{2.45cm}p{2.55cm}p{2.45cm}Y@{}}
\toprule
\textbf{Method} &
\textbf{Memory unit} &
\textbf{Memory control} &
\textbf{Retrieval / use} &
\textbf{Update behavior} &
\textbf{Relation to \method} \\
\midrule

Full context &
Raw conversation history &
No explicit memory mechanism &
Directly places the history in context &
No memory update; limited by context length and cost &
A strong baseline when the full history fits, but it does not learn what should be remembered or compressed. \\

RAG &
Text chunks or document passages &
Designer-defined chunking, indexing, and top-$k$ retrieval &
Retrieves external passages and prepends them to the prompt &
Usually static after indexing; updates require re-indexing &
Provides useful non-parametric access, but memory behavior is fixed by the retrieval pipeline. \\

ReadAgent &
Gist memories and original passages &
Prompted reading policy decides when to summarize and look up details &
Uses gist memories first, then revisits original text when needed &
Constructs task-time gists; no learned long-term memory strategy &
Shows the value of action-based reading and lookup, but does not optimize memory policy from feedback. \\

MemoryBank &
Long-term memories and user profiles &
Framework-defined storage, retrieval, and memory-strength rules &
Retrieves relevant memories using similarity and memory strength &
Updates and strengthens memories over repeated interactions &
Supports persistent personalization, but its memory operations are largely predefined. \\

MemGPT &
Core, recall, and archival memory &
Agent operates within an OS-inspired fixed memory hierarchy &
Page information between active context and archival memory &
Edits core memory and searches archival memory through tool calls &
More adaptive than simple RAG, but still commits to a predefined memory architecture. \\

Mem0 &
Extracted and consolidated facts, optionally graph-structured &
Production memory pipeline controls extraction, consolidation, and retrieval &
Searches stored memories for personalized context &
Continuously consolidates and updates salient facts &
A strong engineered memory layer, but extraction and consolidation remain framework choices rather than learned strategy choices. \\

A-Mem &
Structured notes with tags, links, and contextual attributes &
Agentic note construction following Zettelkasten-style organization &
Retrieves notes together with linked neighboring memories &
Creates, links, and evolves memory notes over time &
Closest to \method in spirit; however, it optimizes within a note-graph design, while \method lets the agent revise the memory strategy itself. \\

Compression &
Compressed summaries or compacted state &
Fixed summarization or compaction policy &
Carries compressed state forward across long histories &
Recompresses windows or states as history grows &
Improves efficiency but can lose task-critical details; \method treats compression as one memory trade-off to be decided. \\

\textbf{\method} &
Agent-written memory workspace plus raw feedback signals &
\textbf{Agent-controlled memory strategy} &
Conditionally writes, retrieves, revises, compresses, or skips memory actions &
\textbf{Revises, merges, deletes, pins, and reorganizes memory based on feedback} &
\textbf{Turns storage, retrieval, update, and compression into editable memory behavior optimized from quality, cost, latency, token, and cache signals.} \\

\bottomrule
\end{tabularx}

\caption{Positioning \method against representative long-context and agent-memory methods. Existing approaches usually instantiate a fixed memory mechanism, such as retrieval, profile memory, hierarchical paging, note graphs, fact extraction, or compression. \method instead exposes these memory behaviors as agent-editable decisions and guides the agent to optimize them using task feedback and efficiency signals.}
\label{tab:memory-components}
\end{table*}

This section describes how we instantiate each memory and context-management baseline in our shared evaluation harness. All methods receive the same \beam conversations and probing questions, and differ only in how they transform the conversation history into the context provided to the answer model.

\paragraph{Shared evaluation protocol.}
All baselines are evaluated in the same harness using identical conversations, probing questions, answer model, and judge. For each scale--method pair, the harness records model outputs, intermediate checkpoints, and request-level usage traces. All model calls are routed through the same official endpoint of OpenAI, ensuring that API usage, prompt caching are measured consistently across methods.

\paragraph{Full context and RAG.}  
The full-context baseline sends the entire rendered conversation history to the answer model, followed by the target question. We report this baseline only when the raw history fits the model and API input limits; larger settings are marked unsupported rather than truncated, so that the baseline reflects true full-context access.
The RAG baseline follows the standard retrieval-augmented generation abstraction~\citep{lewis2020retrieval}. We first flatten the conversation into retrieval chunks. In the default setting, each chunk corresponds to one rendered conversational turn, formatted as \texttt{speaker: utterance}; the chunk metadata stores the original turn index and speaker. We do not merge distant turns or maintain a learned memory state. We embed all chunks using \texttt{text-embedding-3-small} for embeddings, and build an in-memory FAISS vector index. At query time, we embed the question with the same model and retrieve the top-\(k=5\) chunks by cosine similarity. Since all embeddings are normalized, FAISS nearest-neighbor search is equivalent to cosine ranking. The retrieved chunks are concatenated in ranked order and prepended to the answer prompt as non-parametric context.

\paragraph{LoCoMo-style memory.}
The LoCoMo baseline is inspired by the benchmark's dual-memory view of long conversational history \citep{maharana2024locomo}. We convert each session into multiple memory records: a session summary, speaker observations when available, event records, and a fallback session-text memory. At retrieval time, the question embedding ranks all records, but the final context is structured rather than a flat nearest-neighbor list: it contains the latest session summary, relevant session summaries, relevant observations grouped by speaker, and relevant events. This implementation preserves the idea that long-term conversational memory should separate episodic summaries, speaker-specific facts, and event-level memories.

\paragraph{ReadAgent.}
ReadAgent treats long input as a document that should be paginated, gisted, and selectively revisited \citep{lee2024readagent}. We approximate this in the conversational setting by treating each natural session as a page. For each page, we store a short gist and the original page text. The question retrieves relevant gists, then the answer prompt receives both a pagination summary and the selected full pages as an ``interactive lookup'' context. This realization emphasizes ReadAgent's two-stage behavior: lightweight memory first, then targeted access to the underlying page.

\paragraph{MemoryBank.}
MemoryBank models long-term memory with profile information and a forgetting-curve-style retrieval mechanism \citep{zhong2024memorybank}. We instantiate this by storing session summaries, raw turn memories, and speaker-specific observation memories. Each item has a strength value and a temporal distance from the most recent session. Retrieval combines semantic similarity with a retention factor that decays with distance and is moderated by strength. Retrieved memories receive a small strength update. The answer context also includes an overall history summary and personality/profile summary, reflecting MemoryBank's emphasis on persistent user-level memory.

\paragraph{MemGPT.}
MemGPT frames context management as virtual memory, with a small active context and a larger archival store \citep{packer2023memgpt}. Our realization follows this hierarchy. The most recent sessions are placed into working memory together with a compact core-memory block summarizing user/persona information. Older sessions are converted into archival memory records containing a session summary and the session text. At question time, archival memories are retrieved by semantic similarity and paged into the answer context beside the working memory. Thus, the model sees a MemGPT-style split between always-present core/working memory and selectively retrieved archival memory.

\paragraph{A-Mem.}
A-Mem represents memory as linked agentic notes rather than independent chunks \citep{xu2025amem}. We create notes from session summaries and speaker observations; when such structure is unavailable, the full session text becomes a note. Each note stores content, local context, extracted keywords, tags, and links to semantically similar previous notes. Retrieval first selects primary notes by semantic similarity to the question, then expands to linked neighbors. The prompt therefore includes both primary notes and related notes.
\paragraph{Mem0.}
The Mem0 baseline uses the public Mem0 memory SDK as an off-the-shelf long-term memory system. For each source conversation, we create an isolated memory instance backed by a local Qdrant vector store, with a separate directory and collection namespace for that conversation. This prevents memories from different conversations from leaking into one another. Conversation turns are ingested in chronological order in small batches. Each ingested message is rendered with its speaker name and turn index, so that Mem0 can extract memories from the local conversational context rather than from isolated utterances. We use the SDK's standard pipeline: an LLM-based memory extraction step converts raw dialogue into memory candidates, an embedding model encodes the extracted memories, and Qdrant
stores the resulting vectors. 
At evaluation time, each probing question is submitted as a Mem0 search query against the corresponding conversation-specific memory store. We retrieve the top-ranked memories returned by Mem0 and render them as bullet-point context for the answer model. The answer model does not see the full conversation history unless that content has been selected by Mem0's memory extraction and retrieval pipeline. This makes the baseline stronger than simple dense retrieval in that it can store compressed, LLM-extracted facts, but also introduces additional failure modes: relevant information may be omitted during memory extraction, merged incorrectly, or not retrieved at query time. Because Mem0 performs LLM-based extraction, embedding, and vector-store writes during ingestion, it has substantially higher setup overhead than lightweight RAG; we therefore treat ingestion as part of the memory-construction cost and keep each conversation's memory store isolated and reusable across its
 probing questions.

\paragraph{Native compression.}
The compression baseline uses OpenAI Responses native compaction. For histories that exceed the safe active window, the harness rolls over 200K-token windows and carries the compacted state forward with \texttt{previous\_response\_id}. Compression is performed once per conversation, and the compacted state is reused across sequential questions. We record OpenAI-reported cached input tokens from each response.

\section{SelfMem Strategy Iteration Details}
\label{app:selfmem-strategy}

This appendix records the \method strategy-search protocol in a reproducible, prompt-level form. The strategy optimizer used only training-side summaries: aggregate official scores, Pass$_{0.5}$ counts, token-computed costs, LLM/tool request counts, cache statistics, and memory-operation diagnostics. It did not receive held-out questions, held-out answers, rubrics, failed nuggets, source identifiers, or per-question pass/fail labels.

\paragraph{Optimization protocol.}
SelfMem treats memory behavior as a strategy that can be revised by the model itself. At each optimization iteration, the model receives the current procedural strategy together with aggregate training feedback from prior runs. The optimizer is asked to revise only the memory procedure: what to write, what to ignore, how to merge duplicates, when to replace old memory, and how to keep the memory compact enough to be useful under a fixed retrieval budget. The optimizer is explicitly prohibited from writing benchmark facts, answer templates, or conversation-specific content into the strategy. Thus, the search process optimizes the memory-management policy rather than memorizing evaluation answers.

\paragraph{Iteration mechanics.}
Iteration 0 is the first model-generated strategy under the same optimizer constraints, before any subsequent feedback refinement. Later iterations receive only aggregate training summaries from earlier iterations and are asked to produce a revised strategy note. Each candidate strategy is then held fixed while SelfMem processes the training conversations. After the run, only aggregate outcomes are returned to the optimizer. This creates a closed-loop procedure:
\begin{enumerate}
    \item propose a procedural memory strategy;
    \item run SelfMem with that fixed strategy on training conversations;
    \item summarize the run using aggregate score, cost, and memory/tool diagnostics;
    \item revise the strategy without seeing per-question supervision;
    \item repeat for the next optimization iteration.
\end{enumerate}
The held-out conversations are never used to choose an iteration or training-set size. They are used only for final evaluation.

\paragraph{Selected strategy.}
The selected strategy is procedural rather than factual: it changes how the model writes and maintains memory, not what facts it should answer with. The optimizer converged toward a compact-memory policy. The main observation was that larger or more verbose memory was not reliably better; duplicated preference blocks, long project-style summaries, and transient implementation details often consumed context without improving transfer. The selected policy therefore emphasizes precision, stability, and low churn:
\begin{enumerate}
    \item store only stable, high-signal user constraints, preferences, and reusable procedures;
    \item reject domain-specific facts, names, quantities, and scenario-local details;
    \item keep one compact canonical memory object whenever possible;
    \item use short procedure signatures to identify duplicate or near-duplicate memories;
    \item prefer replacement and merging over appending;
    \item keep at most one or two retained memory items unless there are clearly distinct reusable categories;
    \item treat cost as a secondary efficiency signal, never as a reason to drop necessary evidence.
\end{enumerate}

\paragraph{Write and update policy.}
The final strategy uses a strict write gate. A memory is written only when the transcript contains multiple structural signals, such as explicit constraints, precedence rules, required output formats, conflict-resolution behavior, or verification instructions. Weak, transient, or mostly factual candidates are ignored. When a new candidate overlaps with an existing memory, SelfMem replaces or merges the older item instead of appending another copy. This prevents memory growth from diluting the small amount of information that is actually retrieved later.

\paragraph{Read-time implication.}
Because answer-time retrieval is fixed, the strategy focuses on making the stored memory compact and structurally useful. The memory should contain reusable decision procedures, such as how to resolve conflicts or prioritize constraints, rather than long factual summaries. This makes the retrieved memory more likely to influence the model's behavior when only a small memory context is available.

\section{Prompt Templates}
\label{app:prompt-templates}

This appendix provides the prompt-level specification of SelfMem. 
Run-specific values, long JSON payloads, and conversation-specific fields are replaced with placeholders in braces, and line breaks are lightly normalized for readability. 
We include these prompts because SelfMem optimizes memory behavior through prompt-mediated strategy revision rather than through parameter updates. 
Table~\ref{tab:selfmem-prompt-memory-construction} gives the core memory-construction instruction, and Table~\ref{tab:selfmem-prompt-sqlite-addendum} gives the SQLite transcript-mode addendum used when the source conversation is exposed as a read-only chronological table. 
Table~\ref{tab:selfmem-prompt-answer-fixed-evidence} reports the fixed-evidence answering prompt. 
Tables~\ref{tab:selfmem-prompt-local-repair} and~\ref{tab:selfmem-prompt-global-refinement} present the local and global strategy-optimization prompts, respectively. 
Finally, Table~\ref{tab:selfmem-tools} summarizes the internal action space available during strategy construction and memory maintenance.

\definecolor{promptboxbg}{HTML}{F8FAFC}
\definecolor{promptboxborder}{HTML}{2563EB}
\definecolor{promptboxaccent}{HTML}{0F766E}

\begin{table*}[t]
\centering
\begingroup
\setlength{\fboxsep}{10pt}
\setlength{\fboxrule}{0.8pt}
\fcolorbox{promptboxborder}{promptboxbg}{%
\begin{minipage}{0.94\textwidth}
\small
\textbf{\textcolor{promptboxaccent}{Prompt 1: Memory-construction instruction.}}

\medskip
Maintain a model-managed memory workspace for a long-running daily chat. Available tools include \texttt{memory\_read}, \texttt{rag\_search}, \texttt{meta\_log\_read}, \texttt{memory\_change}, \texttt{memory\_review}, and model-created declarative memory tools. Goal: keep memory useful, accurate, source-backed, and cheap for future conversation. Use tools when useful. Always check latest or contradictory exact facts before relying on or saving dates, counts, versions, deadlines, current status, changed values, or user preferences. Prefer atomic exact facts with source turn references when the chat contains dates, counts, versions, deadlines, changed/latest values, or user preferences. Keep broad project profiles separate from exact fact memory. Do not invent facts that are not supported by the chat or tool results.
\end{minipage}%
}
\endgroup
\caption{
Prompt template for SelfMem memory construction. 
The prompt defines the model-managed memory workspace, the available memory tools, and the source-grounding rules used to write accurate, compact, and low-cost future-facing memory.
}
\label{tab:selfmem-prompt-memory-construction}
\end{table*}

\begin{table*}[t]
\centering
\begingroup
\setlength{\fboxsep}{10pt}
\setlength{\fboxrule}{0.8pt}
\fcolorbox{promptboxborder}{promptboxbg}{%
\begin{minipage}{0.94\textwidth}
\small
\textbf{\textcolor{promptboxaccent}{Prompt 2: SQLite transcript-mode addendum.}}

\medskip
The transcript is materialized as one chronological SQLite \texttt{turns} table. Use \texttt{conversation\_manifest} and read-only \texttt{conversation\_sql\_query} to decide how to inspect it. The harness does not split the transcript into pages and does not secretly scan the transcript to build missing-fact lists. Keep memory changes auditable by citing inspected turn ranges. Do not claim coverage for rows you have not actually inspected.
\end{minipage}%
}
\endgroup
\caption{
SQLite transcript-mode addendum. 
The addendum specifies how the model should inspect a chronological transcript table and requires memory edits to remain auditable through cited turn ranges.
}
\label{tab:selfmem-prompt-sqlite-addendum}
\end{table*}

\begin{table*}[t]
\centering
\begingroup
\setlength{\fboxsep}{10pt}
\setlength{\fboxrule}{0.8pt}
\fcolorbox{promptboxborder}{promptboxbg}{%
\begin{minipage}{0.94\textwidth}
\small
\textbf{\textcolor{promptboxaccent}{Prompt 3: Answering with fixed evidence.}}

\medskip
Answer the benchmark question using the memory workspace and retrieved evidence. The memory workspace is a hint/index; retrieved evidence is the authority. Prefer retrieved evidence when it conflicts with memory, and use latest exact values from retrieved evidence for dates, counts, versions, deadlines, and changed facts. For exact-fact questions, compare older and later values in the retrieved evidence; do not stop at the first matching number/date/version. For current/latest/how many/average wording, look for the last matching mention by turn order and prefer the latest confirmed value over older or planned values. Pay attention to wording: if the user asks for first/original values, answer the original value and mention later updates only if useful. If the answer is not supported, say that the available memory/evidence is insufficient. Be concise but include exact dates, versions, names, and constraints when relevant. Inputs: Memory workspace: \{JSON memory view\}; Previous Q/A in this evaluation session: \{last five Q/A pairs\}; Retrieved evidence: \{top retrieved chunks\}; Question: \{question\}.
\end{minipage}%
}
\endgroup
\caption{
Fixed-evidence answering prompt. 
The prompt treats retrieved evidence as authoritative over memory and instructs the model to resolve exact facts, updates, changed values, and insufficient-evidence cases using the retrieved conversation evidence.
}
\label{tab:selfmem-prompt-answer-fixed-evidence}
\end{table*}

\begin{table*}[t]
\centering
\begingroup
\setlength{\fboxsep}{10pt}
\setlength{\fboxrule}{0.8pt}
\fcolorbox{promptboxborder}{promptboxbg}{%
\begin{minipage}{0.94\textwidth}
\small
\textbf{\textcolor{promptboxaccent}{Prompt 4: Local repair.}}

\medskip
Repair the memory and retrieval strategy for training conversation \{conversation\}, attempt \{attempt\}. Target train accuracy: \{target\}\%. The next run will rerun the same training conversation from scratch with your repaired strategy. Learn only from the final official score below; no per-question labels or rubric items are provided. Leakage guardrail: do not request, infer, or mention held-out conversation \{heldout\}, its questions, answers, rubrics, or hidden evaluation signals. Training references may guide strategy repair only; do not store them as future held-out facts. Previous strategy: \{previous strategy\}. Training score feedback: \{aggregate official score JSON\}. Observed memory/tool artifact from the scored attempt: \{artifact summary JSON\}. Write a concise repaired strategy for the next rerun. Improve the general memory and retrieval process based on the score and artifact summary only. Official score/accuracy is the primary objective; \texttt{money\_cost\_usd} and token counts are secondary efficiency signals and tie-breakers. Do not learn a blanket no-retrieval policy just to reduce cost: learn when evidence is needed and prefer targeted RAG/SQL/semantic search over broad dumps. You may reason about when the model should use memory, RAG, SQL, semantic search, or no retrieval, but do not prescribe a fixed tool sequence or hard-code database queries. Keep under 4,000 characters. Return only the repaired strategy note.
\end{minipage}%
}
\endgroup
\caption{
Local strategy-repair prompt. 
The prompt revises the memory and retrieval strategy for a training conversation using only aggregate official score feedback and memory/tool artifact summaries, while enforcing held-out leakage constraints.
}
\label{tab:selfmem-prompt-local-repair}
\end{table*}

\begin{table*}[t]
\centering
\begingroup
\setlength{\fboxsep}{10pt}
\setlength{\fboxrule}{0.8pt}
\fcolorbox{promptboxborder}{promptboxbg}{%
\begin{minipage}{0.94\textwidth}
\small
\textbf{\textcolor{promptboxaccent}{Prompt 5: Global refinement.}}

\medskip
Refine SelfMem's general memory and retrieval strategy from TRAIN score feedback only. Iteration just evaluated: \{iteration\}. Train conversations: \{train conversations\}. Held-out conversation: \{heldout conversation\}. Held-out question count: \{heldout question count\}. Leakage rules: you may inspect train score summaries and train memory/tool diagnostics below; do not request, infer, or mention held-out questions, held-out answers, rubrics, or hidden evaluation signals; do not copy train answers/references as facts to remember for held-out conversations; convert the final scores into a transferable process for unseen conversations. Previous strategy: \{previous strategy\}. Train score feedback: \{score-only feedback JSON\}. Train memory/retrieval/tool diagnostics: \{artifact summaries JSON\}. Write the next candidate strategy as a concise memory-and-retrieval process note. Hard limit: keep it under 4,500 characters. Improve general memory construction and model-chosen retrieval behavior based on the final scores and artifact summaries only. Official score/accuracy is the primary objective; \texttt{money\_cost\_usd} and token counts are secondary efficiency signals and tie-breakers. Do not learn a blanket no-retrieval policy just to reduce cost: learn when evidence is needed and prefer targeted RAG/SQL/semantic search over broad dumps. You may reason about when the model should use memory, RAG, SQL, semantic search, or no retrieval, but do not prescribe a fixed tool sequence, hard-code database queries, or include benchmark answer keys.
\end{minipage}%
}
\endgroup
\caption{
Global strategy-refinement prompt. 
The prompt updates the general SelfMem strategy from training-side aggregate score and diagnostic feedback, without access to held-out questions, answers, rubrics, failed nuggets, or benchmark answer keys.
}
\label{tab:selfmem-prompt-global-refinement}
\end{table*}

\begin{table*}[t]
\centering
\begingroup
\footnotesize
\setlength{\tabcolsep}{3.5pt}
\renewcommand{\arraystretch}{1.12}
\newcommand{\toolcell}[1]{\begin{tabular}[t]{@{}l@{}}#1\end{tabular}}

\begin{tabularx}{\textwidth}{@{}
>{\raggedright\arraybackslash}p{0.15\textwidth}
>{\raggedright\arraybackslash}p{0.25\textwidth}
>{\raggedright\arraybackslash}X
@{}}
\toprule
\textbf{Tool type} & \textbf{Tool} & \textbf{Description} \\
\midrule

Workspace inspection &
\toolcell{
\texttt{memory\_read}\\
\texttt{(target)}
} &
Reads the current model-managed memory workspace, or a specified memory target, before the model decides whether to merge, replace, or leave memory unchanged. \\

Raw-chat retrieval &
\toolcell{
\texttt{rag\_search}\\
\texttt{(query, top\_k)}
} &
Searches the raw conversation log for supporting context during memory construction. It is used to verify what should be written or checked, and does not expose benchmark answers, rubrics, or held-out labels. \\

Run diagnostics &
\toolcell{
\texttt{meta\_log\_read}\\
\texttt{(scope)}
} &
Reads aggregate operational metadata, including token counts, cost, latency, cache statistics, and tool-call counts. These signals help diagnose inefficient memory behavior, while quality remains the primary objective. \\

Memory mutation &
\toolcell{
\texttt{memory\_change}\\
\texttt{(operation, target,}\\
\texttt{content, rationale,}\\
\texttt{metadata)}
} &
Applies a memory operation such as writing, merging, replacing, compacting, indexing, or deleting memory. Each change includes a rationale and optional metadata for auditability. \\

Model-created tool definition &
\toolcell{
\texttt{memory\_tool\_create}\\
\texttt{(name, description,}\\
\texttt{parameters, procedure,}\\
\texttt{rationale, metadata)}
} &
Defines a reusable declarative memory-management tool when primitive operations are insufficient. This supports strategy-level abstraction without hard-coding a fixed memory policy. \\

Model-created tool use &
\toolcell{
\texttt{memory\_tool\_use}\\
\texttt{(name, arguments,}\\
\texttt{target, result\_content,}\\
\texttt{rationale, metadata)}
} &
Applies a previously defined memory-management tool and optionally writes its result into the workspace. This enables the reuse of procedures such as deduplication, compaction, and conflict checking. \\

Self-test creation &
\toolcell{
\texttt{memory\_probe\_create}\\
\texttt{(question, rationale,}\\
\texttt{expected\_memory\_targets,}\\
\texttt{metadata)}
} &
Creates an internal, non-benchmark self-test question to check whether the current memory would support future conversations. The probe is synthetic and does not use held-out evaluation questions or answers. \\

Self-test execution &
\toolcell{
\texttt{memory\_probe\_run}\\
\texttt{(probe\_id, question,}\\
\texttt{query, top\_k,}\\
\texttt{rationale, metadata)}
} &
Runs a self-test against the memory workspace and raw-chat retrieval. The returned evidence is used only to diagnose memory coverage or redundancy, not to tune on benchmark labels. \\

System optimization record &
\toolcell{
\texttt{memory\_system\_optimize}\\
\texttt{(goal, diagnosis,}\\
\texttt{change, validation,}\\
\texttt{rationale, metadata)}
} &
Records a general memory-system improvement or no-op diagnosis based on observed self-test and training-run evidence. This is the main mechanism by which the model revises its memory strategy across iterations. \\

\bottomrule
\end{tabularx}

\caption{
SelfMem internal tools for strategy construction and memory maintenance. 
The tools allow the model to inspect memory state, retrieve raw-chat evidence, read aggregate diagnostics, mutate memory, define reusable memory procedures, run synthetic self-tests, and record strategy-level optimizations. 
They do not expose held-out questions, answers, rubrics, failed nuggets, or per-question labels.
}
\label{tab:selfmem-tools}
\endgroup
\end{table*}

\section{The Use of Large Language Models}  
For this paper, we leveraged GPT-5.4\footnote{\url{https://openai.com/}} and Codex\footnote{\url{https://openai.com/codex/}} to support grammar refinement, LaTeX formatting, and the preparation of figure generation code.  All technical ideas, experimental designs, analyses, conclusions, and writing were
developed and carried out entirely by the authors. The authors have full responsibility for the final
text.  

\end{document}